\documentclass[journal]{IEEEtran}

\usepackage{xcolor}
\usepackage{algorithm}
\usepackage{algpseudocode}
\usepackage{multirow}
\usepackage{amsmath}
\usepackage{mathtools}
\usepackage{graphicx}
\usepackage{xcolor}
\usepackage{enumitem,kantlipsum}
\setlist{nolistsep}

\ifCLASSINFOpdf
\else
\fi
 \usepackage{amssymb}
\usepackage{pifont}
\usepackage{tikz}
\def\cmark{\textcolor{black}{\tikz\fill[scale=0.4](0,.35) -- (.25,0) -- (1,.7) -- (.25,.15) -- cycle;}}
\def\xmark{\textcolor{black}{\ding{53}}}

\usepackage{float,booktabs,array}

\begin{document}
%

\title{Neuroevolution in Deep Neural Networks: \\Current Trends and Future Challenges}
%
%
%

\author{Edgar Galv\'an and Peter Mooney
\thanks{Edgar Galv\'an and Peter Mooney are both with the Naturally Inspired Computation Research Group and with the Department
of Computer Science, Maynooth University, Ireland  e-mails: edgar.galvan@mu.ie and peter.mooney@mu.ie}}

\markboth{}
\markboth{}

%



\maketitle

\begin{abstract}
A variety of methods have been applied to the architectural configuration and learning or training of artificial deep neural networks (DNN). These methods play a crucial role in the success or failure of the DNN for most problems and applications. Evolutionary Algorithms (EAs) are gaining momentum as a computationally feasible method for the automated optimisation and training of DNNs. Neuroevolution is a term which describes these processes of automated configuration and training of DNNs using EAs. While many works exist in the literature, no comprehensive surveys currently exist focusing exclusively on the strengths and limitations of using neuroevolution approaches in DNNs. Prolonged absence of such surveys can lead to a disjointed and fragmented field preventing DNNs researchers potentially adopting neuroevolutionary methods in their own research, resulting in lost opportunities for improving performance and wider application within real-world deep learning problems. This paper presents a comprehensive survey, discussion and evaluation of the state-of-the-art works on using EAs for architectural configuration and training of DNNs. Based on this survey, the paper highlights the most pertinent current issues and challenges in neuroevolution and identifies multiple promising future research directions.  

\end{abstract}


%
\IEEEpeerreviewmaketitle

\section{Introduction}

\label{sec:introduction}

\IEEEPARstart{D}{eep} learning algorithms~\cite{Goodfellow-et-al-2016,10.1162/neco.2006.18.7.1527,DBLP:journals/nature/LeCunBH15}, a subset of machine learning algorithms, are inspired by deep hierarchical structures of human perception as well as production systems. These algorithms have achieved extraordinary results in different exciting areas including computer vision~\cite{7234886},  speech recognition~\cite{6638947,5947494}, board games~\cite{Silver_2016} and video games~\cite{DBLP:journals/corr/MnihKSGAWR13}, to mention a few examples. The design of deep neural networks (DNNs) architectures (along with the optimisation of their hyperparameters) as well as their training plays a crucial part for their success or failure~\cite{LIU201711}.

Architecture search is an area of growing interest as demonstrated by the large number of scientific works published in recent years.  Broadly speaking these works can be classified into one of two categories: evolution-based methods~\cite{Back:1996:EAT:229867,EibenBook2003}, sometimes referred as neuroevolution~\cite{DBLP:journals/evi/FloreanoDM08,784219},  and reinforcement learning (RL) methods~\cite{10.5555/3312046}. Other methods falling outside these two categories have also been proposed in the specialised literature including Monte Carlo-based simulations~\cite{negrinho2017deeparchitect}, random search~\cite{journals/jmlr/BergstraB12} and random search with weights prediction~\cite{DBLP:journals/corr/abs-1708-05344},  hill-climbing~\cite{elsken2017simple}, grid search~\cite{DBLP:conf/bmvc/ZagoruykoK16}, Bayesian optimisation~\cite{10.5555/3042817.3042832,DBLP:conf/nips/KandasamyNSPX18}.

RL architecture-search methods started gaining momentum  thanks to their impressive results~\cite{DBLP:journals/corr/BakerGNR16,DBLP:conf/aaai/CaiCZYW18,DBLP:journals/corr/abs-1712-00559,DBLP:journals/corr/abs-1708-05552,DBLP:journals/corr/ZophL16,DBLP:journals/corr/ZophVSL17}, and more recently, EA architecture-search methods started yielding impressive results in the automatic configuration of DNNs architectures~\cite{Such2017DeepNG,DBLP:conf/iclr/LiuSVFK18,elsken2018neural}. Moreover, it has been reported that neuroevolution requires less computational time compared to RL methods~\cite{DBLP:journals/corr/MnihKSGAWR13,Such2017DeepNG,8742788}.

In their simplest terms a DNN is a feedforward artificial neural network (ANN) with many hidden layers. Each of these layers constitutes a non-linear information processing unit. This simple description encapsulates the incredible capabilities of DNNs. Usually having two or more hidden layers in an ANN qualifies as a DNN. By adding more layers and more units within a layer, a DNN can represent functions of increasing complexity \cite{Goodfellow-et-al-2016}.

Evolutionary Algorithms (EAs)~\cite{Back:1996:EAT:229867,EibenBook2003}, also known as Evolutionary Computation systems, are nature-inspired stochastic techniques that mimic basic principles of life. These \textit{automatic} algorithms have been with us for several decades and are highly popular given that they have proven competitive in the face of challenging problems' features such as discontinuities, multiple local optima, non-linear interactions between variables, among other characteristics~\cite{Eiben:2015:nature}. They have also proven to yield competitive results in multiple real-world problems against other Artificial Intelligent methods as well as results achieved by human experts~\cite{koza_2003,Koza:2010:GPEM}. 

Finding a well-performing architecture is often a very tedious and error-prone process for  Deep Learning  researchers. Indeed, Lindauer and Hutter~\cite{lindauer2019best} remark that there are over $300$ works published in the area of Neural Architecture Search~\cite{lindauer2019best}.  In this work, we focus our attention exclusively in architecture EAs-based search methods in DNNs as well EAs-based approaches in training DNNs. Particularly, this work considers both landmark EAs, such as Genetic Algorithms~\cite{10.5555/531075}, Evolution Strategies~\cite{10.1023/A:1015059928466,10.1007/978-3-642-81283-5_8} and  Genetic Programming~\cite{Koza:1992:GPP:138936}\footnote{Evolutionary Programming~\cite{Fogel1966} is another landmark EA, but to the best of our knowledge, there are no neuroevolution works using this paradigm.} as well as more recent EA variants, such as Differential Evolution~\cite{price_2005}, NeuroEvolution of Augmenting Topologies~\cite{10.1162/106365602320169811}  and Grammatical Evolution~\cite{10.1007/BFb0055930}. Furthermore, we consider the main deep learning architectures, as classified by Liu et al.~\cite{LIU201711},  that have been used in neuroevolution, including  Autoencoders~\cite{10.1145/2598394.2602287}, Convolutional Neural Networks~\cite{10.1145/3065386}, Deep Belief Networks   ~\cite{7508982,DBLP:journals/tnn/0003T0H19} and Restricted Boltzmann Machines~\cite{LIU201711,10.5555/3104322.3104425}. Other deep learning architectures considered in this study include Recurrent Neural Networks~\cite{DBLP:conf/icml/JozefowiczZS15} and Long Short Term Memory~\cite{iet:/content/conferences/10.1049/cp_19991218}.

Previous literature reviews in the area include those conducted by Floreano et al.~\cite{DBLP:journals/evi/FloreanoDM08} and Yao~\cite{784219}, carried out more than one and two decades ago, respectively. More recent works include Stanley et al.~\cite{Stanley2019DesigningNN} and Darwish et al.~\cite{Darwish2019ASO}. The former work explains the influence of modern computational power at scale in allowing the grand ambitions of neuroevolution and deep learning from many years ago to be achieved and fulfilled. The latter work delivers a  broader and high-level introduction and overview of swarm intelligence and EAs in the optimisation of the hyperparameters and architecture for neural networks in the data analytics domain. In contrast to these works, our paper provides a new contribution to complement these studies by concentrating on the details of configuration and design of neuroevolution approaches in deep learning. We carefully consider how EAs approaches are applied in deep learning and in particular their specific configuration for this purpose. 


The goal of our paper is to provide a timely and comprehensive review in neuroevolution in DNNs. The work is aimed at those researchers and practitioners that are either working in the area or are eager to start working in this exciting growing research area. The configuration and design of artificial deep neural networks is error prone, time consuming and difficult. Our paper will highlight the strengths of EAs as a competitive approach to architecture design. We expect this article will attract the attention of researchers in the DL and EA communities to further investigate effective and efficient approaches to addressing new challenges in neuroevolution in DNNs.

Given the specific nature of this paper, an extensive literature review was undertaken. This involved searches in many online databases using a combination of search strategies in order for this review to be methodologically sound. This literature review aims to outline the state-of-the-art and current practice in the domain of neuroevolution by critically evaluating and integrating the findings of all relevant and high-quality individual studies we have found in this area. To establish the extent of existing research and to conduct an exhaustive search representative of all studies that have been conducted on our topic of interest is a major challenge. As the topic of Neuroevolution in Deep Neural Networks straddles several important areas of research in Computer Science: Neural Networks, Machine Learning, and Evolutionary Algorithms there is a need to search widely in order to capture works which appear in one or more of these research areas. We used searches of Google Scholar, IEEE Xplore, ACM Digital Library, ScienceDirect, arXiv, Springer, Citeseer, the archive of Proceedings of Neural Information Processing Systems, and the archive of Proceedings of International Conference on Machine Learning. We strongly believe our paper outlines our estimation of the state-of-the-art in neuroevolution in DNNs at this point in time. 



The rest of the paper is organised as follows. Section~\ref{sec:background} provides some background  to DL and  EAs. Section~\ref{sec:architectures} discusses how the architectures of DNNs can be evolved efficiently using EA approaches. This moves onto a discussion in Section~\ref{section:learningDNNsThroEAs} on the training of DNNs with EAs.  Section~\ref{sec:challenges}  sets out some of the major challenges and fertile avenues for future work. Finally, the paper closes with some concluding remarks in Section~\ref{sec:conclusions}.

\section{Background}
\label{sec:background}


\subsection{Deep Neural Networks}

The concept of deep learning originated from the study on artificial neural networks (ANNs). An ANN consists of multiple, simple, connected units denominated neurons, each producing a sequence of real-valued activations and, by carefully adjusting the weights, the ANNs can behave as desired. Depending on the problems and the way the neurons are connected, the process of training an ANN may require  ``long casual chains of computational stages"~\cite{SCHMIDHUBER201585}. Deep learning emerged as a concept from works such as Hinton et al.~\cite{10.1162/neco.2006.18.7.1527} and has subsequently became a very active research area~\cite{LIU201711}. A deep learning algorithm is a class of machine learning algorithms using multiple layers to progressively extract higher level features from the raw data input. The term deep then refers specifically to the number of layers through which the raw data is transformed. In deep learning, each subsequent level attempts to learn in order to transform input data into a progressively more abstract and composite representation. Neuroevolution in DNNs has been applied to the development of a wide range of ANNs including, but not limited to, Convolutional Neural networks, Autoencoders, Deep Belief Networks and Recurrent Neural Networks. In the next sections, we summarise these. 

\subsubsection{ Deep Learning Architecture: Convolutional Neural Networks (CNNs)}

CNNs have shown impressive performance in processing  data with a grid-like topology.  The deep network consists of a set of layers each containing one or more planes. Each unit in a plane  receives input from a neighbourhood in the planes of the previous layer. This idea of connecting units to receptive fields dates back to the 1960s with the perceptron and the animal visual cortex organisation discovered by Hubel and Wiesel~\cite{doi:10.1113/jphysiol.1962.sp006837}. The input, such as an image, is convolved with trainable kernels or filters at all offsets to produce feature maps. These filters include a layer of connection weights. Usually, four pixels in a feature map form a group and this is passed through a function, such as sigmoid function or hyperbolic tangent function. These pixels produce additional feature maps in a layer.  $n$ planes are normally used in each layer so that $n$ features can be detected. These layers are called convolutional layers. Once a feature is detected, its exact location is less important and convolutional layers are followed by another layer in charge of performing local averaging and sub-sampling operation. Due to the high dimensionality of the inputs, a CNN classifier may cause overfitting. This problem is addressed by using a pooling process, also called sub-sampling or down-sampling, that reduces the overall size of the signal. Normally, the CNN is trained with the usual backpropagation gradient-descent procedure proposed by Lecun et al.~\cite{Lecun98gradient-basedlearning}.

The learning attractive process of a CNN is determined by three key elements: (i) sparse interaction that reduces the computational processing with kernels that are smaller than the inputs, (ii) parameter sharing that refers to learn one set of parameters instead of learning one set at each location, and finally, (iii) equivariance representation that means that whenever the input changes, the output changes in the same manner~\cite{Goodfellow-et-al-2016}. CNNs were the first successful \textit{deep learning architectures} applied to face detection, handwriting recognition, image classification, speech recognition, natural language processing and recommender systems.

The evolution of these CNN architectures has been slow but remarkable. For  example, LeNet~\cite{Lecun98gradient-basedlearning} proposed in the late 1990s and AlexNet~\cite{10.1145/3065386}, proposed a decade later, are very similar with two and five convolutional layers, respectively. Moreover, they also used kernels with large receptive fields in the layer close to the input and smaller filters closer to the output. A major difference is that the latter used rectified linear units as activation function, which became a standard in neuroevolution in designing CNNs.  Since 2012, the use of novel and deeper models took off. For example, in 2014, Simonyan and Zisserman~\cite{DBLP:journals/corr/SimonyanZ14a} won the Imagenet challenge with their proposed 19-layer model known as VGG19. Other networks have been proposed that not only are deeper but use more complex building blocks. For example, in 2015, Szegedy et al.~\cite{DBLP:conf/cvpr/SzegedyLJSRAEVR15} proposed GoogLeNet, also known as Inception, which is a 22-layer network that used inception blocks. In 2015, the ResNet architecture, consisting of the so-called ResNet blocks, proposed by He et al.~\cite{DBLP:conf/cvpr/HeZRS16} won the ImageNet challenge. Moreover, multiple CNNs variants have been proposed such as combining convolutions with an autoencoder~\cite{5459469}, using RBMs in a CNN~\cite{Desjardins-2008}, to mention a few examples. A description of the variants of this network can be found in~\cite{LIU201711}.

\subsubsection{ Deep Learning Architecture: Autoencoders (AEs)}
\label{section:autoencoders}

Autoencoders are simple learning circuits which are designed to transform inputs into outputs with the minimum amount of distortion. An autoencoder consists of a combination of an encoder function and a decoder function. The encoder function converts the input data into a different representation and then the decode function converts the new representation back to the original form. AEs attempt to preserve as much information as possible and they provide range-bounded outputs which make them suitable for data pre-processing and iterative architectures such as DNNs~\cite{7273701}. Despite this work appearing in $2020$, the authors suggest that there is ``still relatively little work exploring the application (of evolutionary approaches to neural architecture search) to autoencoders. In $2012$ Baldi~\cite{pmlr-v27-baldi12a} argued that while autoencoders ``taken center stage in the deep architecture approach'' there was still very little theoretical understanding of autoencoders with deep architectures to date. Choosing an appropriate autoencoder architecture  in order to process a specific dataset will mean that the autoencoder is capable of learning the optimal representation of the data~\cite{CharteEvoAA2020}. Encoding autoencoders within a chromosone representation means that such an approach could be broad enough to consider most autoencoder variations~\cite{CharteEvoAA2020}. As an unsupervised feature learning approach, autoencoders attempt to learn a compact representation of the input data whilst retaining the most important information of the representation. This representation is expected to completely reconstruct the original input. This makes initialisation of the autoencoder critical~\cite{7293666Autoencoders}. Whilst autoencoders can induce very helpful and useful representations of the input data they are only capable of handling a single sample and are not capable of modelling the relationship between pairs of samples in the input data.

\subsubsection{Deep Learning Architecture: Deep Belief Networks (DBNs)}

Deep Belief Networks  can be implemented in a number of ways including Restricted Boltzmann Machines (see Section~\ref{subsection:othernetworks}) and Autoencoders (see Section~\ref{section:autoencoders}). DBNs are well suited to the problem of feature extraction and have drawn ``tremendous attention recently''~\cite{DBLP:journals/tnn/0003T0H19}. DBNs, like other traditional classifiers, have a very large number of parameters and require a great deal of training time~\cite{7744376}. When Restricted Boltzmann Machines (RBMs) are stacked together they are considered to be a DBN. The fundamental building blocks of a DBN are RBMs consisting of one visible layer and one hidden layer. When DBNs are applied to classification problems the feature vectors from data samples are used to set the values of the states of the visible variables of the lower layer of the DBN. Then the DBN is trained to generate a probability distribution over all possible labels of the input data. They offer a good solution to learn hierarchical feature representations from data.

\subsubsection{Deep Learning Architecture: Other network types}
\label{subsection:othernetworks}
In this subsection we introduce some of other popular and well-studied network architectures namely: Recurrent Neural Networks (RNNs),  Restricted Boltzmann Machines (RBMs), and Long Short Term Memory (LSTM). 

\textbf{RNNs}: In the case of CNNs input is a fixed-length vector and eventually produce a fixed-length vector as output. The number of layers in the CNN determine the amount of computational steps required. Recurrent Neural Networks (RNNs) are more flexible and they allow operation across a sequence of vectors. The connections between the units in the network form a directed cycle and this creates an internal state of the network allowing to exhibit dynamic temporal behaviour. This internal hidden state allows the RNN to store a lot of information about the past efficiently. RNNs are well suited to sequential data prediction and this has seen them being applied to areas such as statistical language modelling and time-series prediction. However, the computational power of RNNs make them very difficult to train. The principal reasons for this difficulty are mainly due to the exploding and the vanishing gradient problems~\cite{DBLP:conf/icml/JozefowiczZS15}. In theory RNNs can make use of information in arbitrarily long sequences  but in reality they are limited to considering  look-back at only a few steps.

\textbf{RBMs:} A Restricted Boltzmann Machine (RBM) is a network of symmetrically connected neuron like units which are designed to make stochastic decisions about whether to be on or off. They are an energy-based neural network. In a  RBM there are no connections between the hidden units and multiple hidden layers. Learning occurs by considering the hidden activities of a single RBM as the data for training a higher-level RBM~\cite{pmlr-v5-salakhutdinov09a}. There is no communication or connection between layers  and this is where the \textit{restriction} is introduced to a Boltzmann machine. The RBMs are probabilistic models using a layer of hidden binary variables or units to model the distribution of a visible layer of variables. RBMs have been successfully applied to problems involving high dimensional data such as images and text~\cite{JMLR:v13:larochelle12a}. As outlined by Fischer and Igel~\cite{10.1007/978-3-642-33275-3_2}, RBMs have been the subject of recent research after being proposed as building blocks of multi-layer learning architectures or DBNs. The concept here is that hidden neurons extract relevant features from the data observations. These features can then serve as input to another RBM. By this so-called stacking of RBMs in this fashion way, a network can learn features from features with the goal of arriving at a high level representation~\cite{5947494}.

\textbf{LSTM:} 
Long-short-term memory (LSTM) networks are a special type of recurrent neural networks capable of learning long-term dependencies. They work incredibly well on a large variety of problems and are currently widely used. LSTMs are specifically designed to avoid the problem of long-term dependencies. The basic unit within the hidden layer of an LSTM network is called a memory block containing one or more memory cells and a pair of adaptive, multiplicative gating units which gate input and output to all cells in the block~\cite{iet:/content/conferences/10.1049/cp_19991218}. In LSTM networks, it was possible to circumvent the problem of the vanishing error gradients in the network training process by method of error back propagation. An LSTM network is usually controlled by recurrent gates called “forgetting” gates. Errors are propagated back in time through a potentially unlimited number of virtual layers. In this way, learning takes place in LSTM, while preserving the memory of thousands and even millions of time intervals in the past. Network topologies such as LSTM can be developed in accordance with the specifics of the task. Recurrent neural networks (RNNs) with long short-term memory (LSTM) have emerged as an effective and scalable model for several learning problems related to sequential data~\cite{7508408}. Gers and Schmidhuber~\cite{963769} showed that standard RNNs fail to learn in the presence of time lags exceeding as few as five to ten discrete-time steps between relevant input events and target signals. LSTM are not affected by this problem and are capable of dealing with minimal time lags in excess of 1000 discrete-time steps. In studies such as those by Gers and Schmidhuber~\cite{963769},  LSTM clearly outperforms previous RNNs not only on regular language benchmarks (according to previous research) but also on context-free languages benchmarks.


\subsection{Evolutionary Algorithms}


\begin{algorithm*}
  \caption{A common EA process for network design. Adapted from~\cite{8237416}}
         
\begin{algorithmic}[1]
\State \noindent \textbf{Input:} the reference dataset $D$, the number of generations $T$, the number of individuals in each generation $N$, the mutation and crossover probabilities $P_m$ and $P_c$;
\State \noindent \textbf{Initialisation:} generating a set of randomised individuals $\{{\mathbb{M}_0,n\}}^N_{n=1}$, and computing their recognition accuracies;


\For {$t = 1, 2, \cdots, T$}
\State \textbf{Selection:} producing a new generation $\{\mathbb{M}'_{t,n}\}^N_{n=1}$ with a Russian roulette process on $\{\mathbb{M}_{t-1,n}\}^N_{n=1}$;
\State \textbf{Crossover:} for each pair $(\{\mathbb{M}_{t,2n-1},\mathbb{M}_{t,2n})\}^{[N/2]}_{n=1}$, performing crossover with probability $P_c$;
\State \textbf{Mutation:} for each non-crossover individual $\{\mathbb{M}_{t,n}\}^N_{n=1}$, performing mutation with probability $P_m$;
\State \textbf{Fitness evaluation:} computing the fitness (e.g., recognition accuracy) for each new individual $\{\mathbb{M}_{t,n}\}^N_{n=1}$;
\EndFor
                     
\\
\noindent \textbf{Output:} a set of individuals in the final generation  $\{\mathbb{M}_T,n\}^N_{n=1}$ with their fitness values. 
\end{algorithmic}
\label{alg:EAs}
\end{algorithm*}

Evolutionary Algorithms (EAs)~\cite{Back:1996:EAT:229867,EibenBook2003} refer to a set of stochastic optimisation bio-inspired algorithms that use evolutionary principles to build robust adaptive systems. The field has its origins in four landmark evolutionary methods: Genetic Algorithms~\cite{10.5555/531075,DBLP:books/aw/Goldberg89}, Evolution Strategies~\cite{Rechenberg10.1007/978-3-642-83814-9_6,10.5555/539468}, Evolutionary Programming~\cite{Fogel1966} and Genetic Programming~\cite{Koza:1992:GPP:138936}. The key element of these algorithms is undoubtedly flexibility allowing the practitioner to use elements from two or more different EAs techniques. This is the reason why the boundaries between these approaches are no longer distinct allowing to have a more holistic EA framework~\cite{10.5555/1841436} via~\cite{FLEMING20021223}.

EAs work with a population of $\mu$-\textit{encoded} potential solutions to a particular problem. Each potential solution, commonly known as individual, represents a point in the search space, where the optimum solution lies. The population is evolved by means of genetic operators, over a number of generations, to produce better results to the problem.  Each member of the population is evaluated using a fitness function to determine how good or bad the potential solution is in the problem at hand. The fitness value assigned to each individual in the population probabilistically determines how successful the individual will be at propagating (part of) its code to further generations. Better performing solutions will be assigned higher values (for maximisation problems) or lower values (for minimisation problems).  

The evolutionary process is carried out by using genetic operators. Most EAs include operators that select individuals for reproduction, generate new individuals based on the selected individuals and ultimately determine the composition of the individuals in the population at the following generation. Selection, crossover and mutation are key genetic operators used in most EAs paradigms. 
The selection operator is in charge of choosing one or more individuals from the population based on their fitness values. Multiple selection operators have been proposed. One of the most popular selection operators is tournament selection for its simplicity. The idea is to select the best individual from a pool, normally of size = $[2-7]$, from the population. The stochastic crossover, also known as recombination, operator exchanges material normally from two selected individuals. This operator is in charge of exploiting the search space. The stochastic mutation operator makes random changes to the genes of the individual. This operator is in charge of exploring the search space. The mutation operator is important to guarantee diversity in the population as well as recovering genetic material lost during evolution. 

The evolutionary  process explained before is repeated until a condition is met. Normally, until a maximum number of generations has been executed. The population in the last generation  is the result of exploring and exploiting the search space over a number of generations. It contains the best evolved potential solutions to the problem and may also represent the global optimum solution.  Alg.~\ref{alg:EAs} shows the general steps of a EA for a deep CNN network design.

\subsubsection{Evolutionary Algorithm: Genetic Algorithms (GAs)} This EA was introduced by Holland~\cite{10.5555/531075} in the 1970s and highly popularised by Goldberg~\cite{DBLP:books/aw/Goldberg89}. This was due to the fact of achieving extraordinary results as well as reaching multiple research communities including machine learning and neural networks. GAs were frequently described as function optimisers, but now the tendency is to consider GAs as search algorithms able to find near-optimal solutions. Multiple forms of GAs have been proposed in the specialised literature. The bitstring fixed-length representation is one of the most predominant encodings used in GAs. Crossover, as the main genetic operator, and mutation as the secondary operator, reproduce offspring over evolutionary search. 

\subsubsection{Evolutionary Algorithm: Genetic Programming (GP)} This EA is a subclass of GAs and was popularised by Koza~\cite{Koza:1992:GPP:138936} in the 1990s. GP is a form of automated programming. Individuals are randomly created by using  functional and terminal sets, that are required to solve the problem at hand. Even though multiple types of GP have been proposed in the specialised literature, the typical tree-like structure is the predominant form of GP in EAs. Cartesian GP~\cite{Miller2011} is another form of GP, which has been used in neuroevolution in DNNs~\cite{DBLP:conf/icml/SuganumaOO18,ijcai2018-755}.

\subsubsection{Evolutionary Algorithm: Evolution Strategies (ES)} These EAs were introduced in the 1960s by Rechenberg~\cite{Rechenberg10.1007/978-3-642-83814-9_6} and Schwefel~\cite{10.5555/539468}. ES are generally applied to real-valued representations of optimisation problems. In ES, mutation is the main operator whereas crossover is the secondary, optional, operator. Historically, there were two basic forms of ES, known as the ($\mu,\lambda$)-ES and the ($\mu+\lambda$)-ES. $\mu$ refers to the size of the parent population, whereas $\lambda$ refers to the number of offspring that are produced in the following generation before selection is applied. In the former ES, the offspring replace the parents whereas in the latter form of ES, selection is applied to both offspring and parents to form the population in the following generation. Nowadays, the Covariance Matrix Adaptation-ES, proposed in the 1990s by Hansen~\cite{6790790,542381,10.1162/106365601750190398}, is the state of the art ES that adapts the full covariance matrix of a normal search (mutation) distribution.

\subsubsection{Evolutionary Algorithm: Evolutionary Programming (EP)} These EAs were proposed in the 1960s by Fogel, Owens and Walsh~\cite{Fogel1966} and very little differences are observed between ES and EP. The main difference, however, between these two EAs paradigms is the lack of use of crossover in EP whereas this genetic operator is secondary, and rarely used, in ES. Another difference is that in EP, normally $M$ parents produce $M$ offspring, whereas in ES the number of offspring produced by genetic operators is higher than their parents.

\subsubsection{Evolutionary Algorithm: Others}

Multiple evolutionary-based algorithms have been proposed in the specialised literature. Relevant to this work are  Differential Evolution (DE), Grammatical Evolution (GE) as well as NeuroEvolution of Augmenting Topologies (NEAT).

\textbf{DE:} Differential Evolution (DE) was proposed by Price and Storn~\cite{price_2005} in the 1990s. The popularity of this EA is due to the fact that has proven to be highly efficient in continuous search spaces and it is often reported to be more robust as well as achieving a faster convergence speed compared to other optimisation methods~\cite{4358759}. Unlike traditional EAs, the DE-variants perturb the population members with the scaled differences of randomly selected and distinct population members.

\textbf{GE:} Grammatical Evolution (GE) is a  grammar-based EA proposed by Ryan et al.~\cite{10.1007/BFb0055930} in the 1990s. A genotype-phenotype mapping process is used to generate (genetic) programs  by using a binary string to select production rules in a Backus-Naur form grammar definition. GE can be seen as a special form of GP, where one of the main differences is that unlike GP, GE does not perform the evolutionary process on the programs themselves.

\textbf{NEAT:} NeuroEvolution of Augmenting Topologies (NEAT) is a form of EA  proposed by Stanley and  Miikkulainen~\cite{10.1162/106365602320169811} in the 2000s. NEAT is a technique for evolving neural networks.    Three elements are crucial for NEAT to work: (i) historical marking, that allows solutions to be crossed over, (ii) speciation, that allows for defining niches and (iii)  starting from minimal structure, that allows to incrementally find better solutions.

\section{Evolving DNNs Architectures Through Evolutionary Algorithms}
\label{sec:architectures}

\subsection{The Motivation}{}

In recent years, there has been a surge of interest in methods for neural architecture search. Broadly, they can be categorised in one of two areas: evolutionary algorithms or reinforcement learning. Recently, EAs have stated gaining momentum  for designing deep neural networks architectures~\cite{elsken2018neural,DBLP:journals/evi/FloreanoDM08,Kitano1990DesigningNN,DBLP:conf/iclr/LiuSVFK18,DBLP:journals/corr/MiikkulainenLMR17,DBLP:conf/aaai/RealAHL19,10.5555/3305890.3305981,10.1162/106365602320169811,8237416}. The popularity of these algorithms is due to the fact that they are gradient-free, population-based methods that offer a parallelised mechanism to simultaneously explore multiple areas of the search space while at the same time offering a mechanism to escape from local optima. Moreover, the fact that the algorithm is inherently suited to parallelisation means that more potential solutions can be simultaneously computed within acceptable wall-clock time. Steady increase in computing power, including graphics processing units with thousands of cores, will contribute to speed up the computational calculations of population-based EAs.

\subsection{The Critique}

Despite the popularity of EAs for designing deep neural network architectures, they have also been criticised in the light of being slow learners as well as being computationally expensive to evaluate~\cite{Eiben:2015:nature}. For example, when using a small population-based EA of 20 individuals (potential solutions) and using a training set of 50,000 samples, one generation alone (of hundreds,  thousands or millions of generations) will require one million evaluations through the use of a fitness function. 

\subsection{Deep Learning Architecture: Convolutional Neural Networks}

\begin{figure*}[tbh!]
  \centering
      \includegraphics[width=0.85\textwidth]{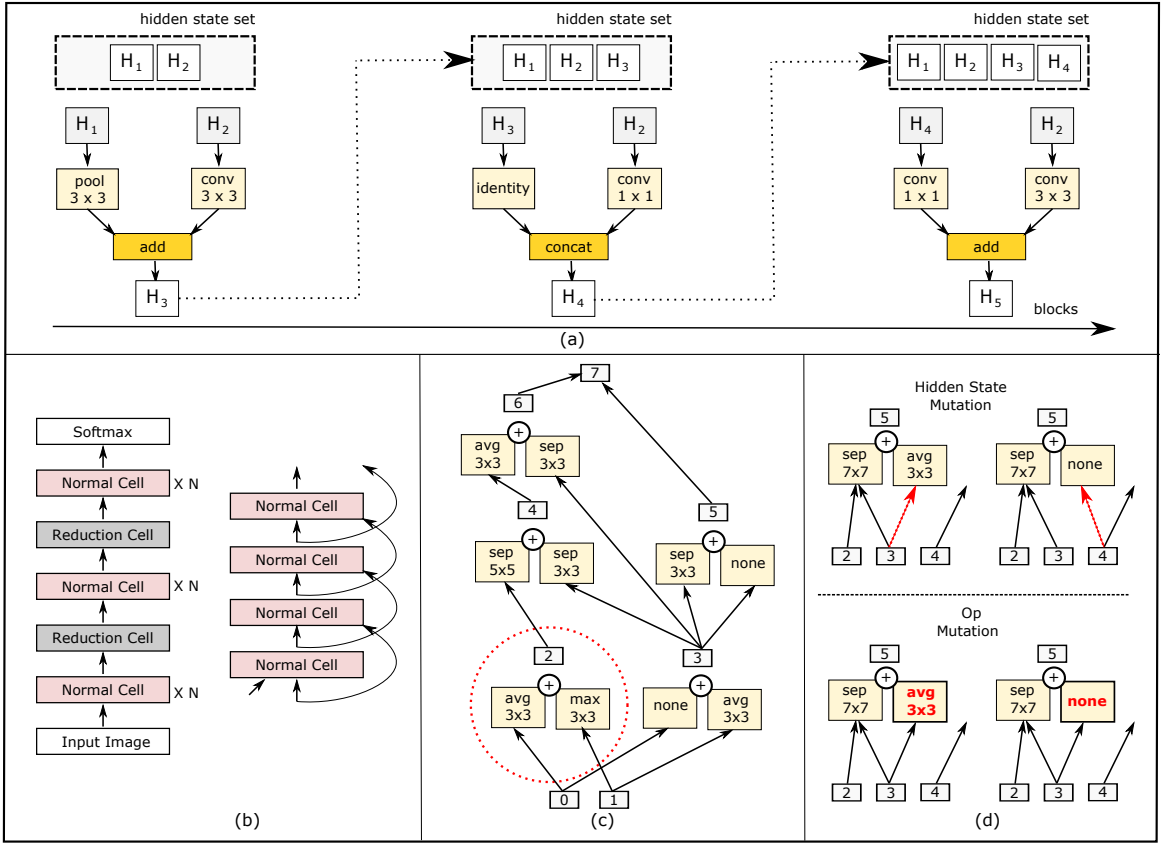}
\caption{\textbf{(a)} NASNet Search Space~\cite{DBLP:journals/corr/ZophVSL17}.
\textbf{(b)} Scalable architecture for image classification consist of two repeated motifs termed Normal Cell and Reduction Cell. Left: the full outer structure (omitting skip inputs for clarity) and Right: detailed view with the skip inputs. \textbf{(c)} Example of a cell, where the dotted red circle demarcates a pairwise combination. \textbf{(d)} Examples of how the mutations are used. \textbf{(a)} and \textbf{(b -- d)} redrawn from Zoph et al.~\cite{DBLP:journals/corr/ZophVSL17} and Real et al.~\cite{DBLP:conf/aaai/RealAHL19}, respectively.}
\label{fig:NASNet}
\end{figure*}

 Agapitos et al.~\cite{DBLP:conf/cec/AgapitosONFKBC15} used a tree-based GP system to evolve a hierarchical feature construction as well as a classification system with feedforward processing.  Inspired by the work carried out by Bengio et al.~\cite{NIPS2006_3048} where they empirically demonstrated, using DBN, that greedy-unsupervised layer-training strategy helps to optimise deep  networks, Agapitos et al. trained a new layer of CNN of image transformation at the time with the goal to learn a stack of gradually better representations. The authors attained good results using the MNIST dataset. Dufourq and Bassett~\cite{8261132} used a GA to evolve CNN architectures. The encoding used by the authors also allowed them to evolve the learning rate denoting the value which is applied during the training optimisation. They used different operations and sizes of filters including one and two-dimension convolution, one and two-dimension max pooling,  dropout, among others. They tested their approach in both image classification tasks  as well as in sentiment analysis tasks. Chromosomes using two-dimensional convolution were penalised for a text sentiment problem.  The authors used a fitness function with two aggregate elements, one denoting the accuracy and the other the complexity of the solution captured by the number of parameters used by a chromosome. The authors reported competitive results compared to state-of-the-art algorithms on the balanced-based and digit-based EMNIST dataset as well as in the Fashion dataset.   Desell~\cite{DBLP:conf/gecco/Desell17} proposed an algorithm based on NEAT~\cite{10.1162/106365602320169811} to evolve CNN architectures. Desell carried out some modifications to the NEAT algorithm to evolve CNN architectures through selection, crossover and mutation. Whereas all operators played an important role to produce well-behaved CNNs, it was interesting to see how the use of mutation, involving 7 types of operations, seemed to be crucial to the competitive results reported on the MNIST dataset. 

Zoph et al.~\cite{DBLP:journals/corr/ZophVSL17} proposed NASNet search space defined by a predetermined outer structure depicted in Fig.~\ref{fig:NASNet} with the ultimate goal of enabling transferability.  This structure is composed of convolutional cells, called normal cell (coloured in pink in Fig.~\ref{fig:NASNet} (a)) and reduction cell (coloured in grey), repeated many times. The former type of cells returns a feature map of the same dimensions whereas the latter returns a feature map where its height and width is reduced by a factor of two.  All cells of the same type are constraint to have the same architecture and it was found beneficial that architectures of normal cells were different to the architectures of reduced cells. The goal of their architecture search process was to discover the architectures of these two types of cells, an example of this is denoted in Fig.~\ref{fig:NASNet} (b).  In 2019, Real et al.~\cite{DBLP:conf/aaai/RealAHL19} proposed AmoebaNet-A to evolve an image classifier achieving superior accuracy over hand-designed methods for the first time. The authors used a population-based EA with each fixed length member encoding the architecture of CNNs. To do so, they used the NASNet search space~\cite{DBLP:journals/corr/ZophVSL17}. The goal of their EA-based approach was to discover the architectures of the normal cells and the reductions cells as depicted in Fig.~\ref{fig:NASNet} (a). Real et al. used a modified version of tournament selection and two types of mutation as the two genetic operators in charge of driving evolution. Tournament selection (see Section~\ref{sec:background} to read about how this works) was modified so that the newest genotypes were chosen over older genotypes.  The mutation operator involved one of two operations taking place once for each individual: the hidden state mutation and the op mutation. To execute any of these types of mutation, first a random cell is chosen, then a pairwise combination is stochastically selected (see Fig.~\ref{fig:NASNet} (c)), and finally, one of these two pairs is picked randomly. This hidden state is replaced with another hidden state with the constraint that no loop is formed. The op mutation  differs only in modifying the operation used within the selected hidden state. Fig.~\ref{fig:NASNet} (d) shows how these two mutation operations work. The authors used the CIFAR-10 dataset to test their proposed AmoebaNet-A and compared it against a reinforcement learning-based method and random search, achieving better accuracy results as well as reducing the computational time required by their algorithm compared to the other two methods. Moreover, the authors used the fittest chromosome found by their algorithm and retrained it using the Imagenet dataset. With this, they also reported encouraging accuracy results compared with other architecture search methods. 

In a different constraint setting, Xie et al.~\cite{8237416} proposed Genetic CNN to automatically learn the structures of CNNs with a limited number of layers as well as limited sizes and operations of convolutional filters, to mention a few constraints adopted by them. The authors adopted a GA with binary fixed-length representation to represent parts of evolved network. In their studies, each network is composed by various stages and each of these is composed of nodes that represent convolutional operations.  The binary encoding adopted by Xie et al. represents the connection between a number of ordered nodes. This representation allowed the authors to use crossover, along with roulette selection and mutation. They defined a stage as the minimal unit to apply crossover and mutation. This allowed them to maintain the ordered nodes defined in the genotype and produce valid potential solutions only. Even with the restrictions adopted in their work, the authors achieved competitive accuracy results using the CIFAR-10 and MNIST datasets. They also demonstrated how their approach can be generalised by using the learned architecture on the ILSVRC2012 large-scale dataset. This was achieved  because their approach was able to produce chain-shaped networks such as AlexNet~\cite{10.1145/3065386}, VGGNet~\cite{DBLP:journals/corr/SimonyanZ14a}, multiple-path networks such as GoogLeNet~\cite{DBLP:conf/cvpr/SzegedyLJSRAEVR15} and highway networks such as Deep ResNet~\cite{DBLP:conf/cvpr/HeZRS16}, which have been reported to be beneficial when applied to computer vision problems.

 Real et al.~\cite{10.5555/3305890.3305981} used an EA to automatically optimise CNN architectures. Individual architectures are encoded as a graph, where the vertices represent rank-3 tensors: two of these represent the spatial coordinates of the image and the third is the number of channels. Activation functions, such as batch-normalisation~\cite{DBLP:conf/icml/IoffeS15} with rectified linear units (ReLus) or plain linear units, are applied to the vertices. The authors primarily used 11 types of mutations falling into one of three different categories including inserting layers, removing layers as well as using a mutation to modify layers parameters. Although they also conducted studies using three forms of crossover, Real et al. indicated that none of these improved the results yielded by mutation operators. The authors also indicated that at the beginning of the search, the EA was susceptible to becoming trapped at local optima. To this end, they applied five mutations per reproduction and decreased this to one at a later stage during evolution. Real et al. used back-propagation to optimise the weights of the CNNs. Because training a large model is incredible slow within their evolutionary setting, the authors partly addressed this by allowing children to inherit their parents' weights when layers had matching shapes. They also used a high-computational setting (250 computers) to carry out their experiments.  In their results, the authors reported competitive average accuracy results over five independent runs in the CIFAR-10 and CIFAR-100  datasets compared to state-of-the-art algorithms including ResNet~\cite{DBLP:conf/cvpr/HeZRS16} and DenseNet~\cite{8099726}. 

\begin{figure*}[tbh!]
  \centering
      \includegraphics[width=0.85\textwidth]{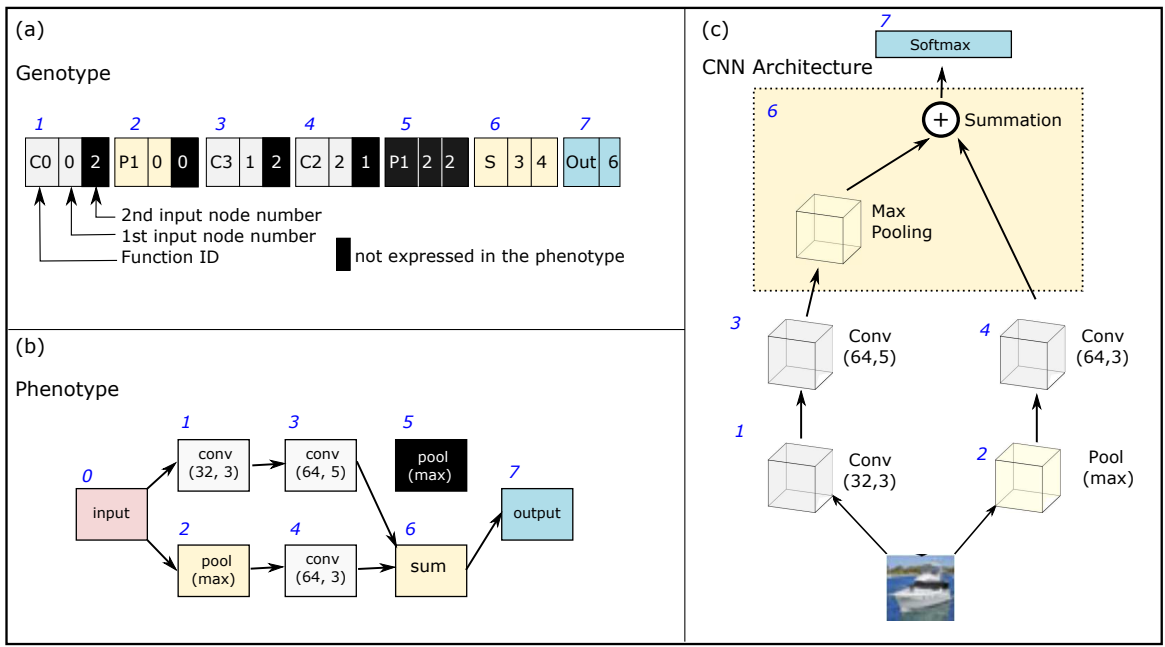}
\caption{\textbf{(a)} Genetic representation of a cartesian genetic programming individual encoding a CNN architecture. \textbf{(b)} The phenotypic representation. \textbf{(c)} The CNN architecture defined by (a). Notice that the Gene No. 5, coloured with a black background  in the genotype (a) is not expressed in the phenotype. The summation node in (c), highlighted with a light yellow background, performs max pooling to the left-hand side of the input (Node no. 3) to get the same input tensor sizes.  Redrawn from Suganuma et al.~\cite{ijcai2018-755}.}
\label{fig:cartesianCNN}
\end{figure*}

 Suganuma et al.~\cite{ijcai2018-755} used Cartesian GP~\cite{Miller2011} (CGP) to automatically design CNN architectures. The genotype encodes information on the type and connections of the nodes. Fig.~\ref{fig:cartesianCNN} (a) depicts this idea. 
 These types include ConvBlock,  ResBlock, max pooling, average pooling, summation and concatenation. ConvBlock consists of standard convolution processing followed by batch normalisation and ReLU~\cite{10.5555/3104322.3104425} whereas ResBlock is a ConvBlock followed by tensor summation.  The CGP encoding scheme represents the program as a directed acyclic graph in a two-dimensional grid of $N_r$ rows by $N_c$ columns. Fig.~\ref{fig:cartesianCNN} (b) provides an example of the phenotype, obtained from Fig.~\ref{fig:cartesianCNN} (a), in the case of a grid defined by $N_r=2$ by $N_c=3$. The corresponding CNN architecture is depicted in Fig.~\ref{fig:cartesianCNN} (c). In their experiments, the authors used the CIFAR-10 dataset as well as a portion of this. As evaluating each of the CGP individuals is  expensive, they adopted a simple  (1+$\lambda$) ES (see Section~\ref{sec:background}).  The authors' approach achieved competitive results compared with well-known methods including ResNet~\cite{DBLP:conf/cvpr/HeZRS16}. It is interesting to see how the authors reported encouraging results using CGP to automatically configure CNN architectures regardless of the sample size used in their work. For example, the CNN architecture produced by CGP for the small dataset scenario is wider compared to the architecture yield by CGP for the standard scenario.

 Assun{\c{c}}{\~{a}}o et al.~\cite{10.1007/978-3-319-77553-1_2,DBLP:journals/gpem/AssuncaoLMR19} proposed DENSER, an hybrid mechanism of GAs and (Dynamic Structured) Grammatical Evolution (GE)~\cite{10.1007/BFb0055930}, to  evolve DNNs architectures.  The outer layer of their proposed  approach is in charge of encoding the macro structure of the DNNs evolved by means of GAs. Dynamic Structured GE is in charge of the inner layer that encodes the parameters of the DNNs in a backus-naur form. The authors used the typical genetic operators, including selection, two forms of crossovers (one-point and bit-mask)  and three types of mutations (add, replicate and remove unit) in the outer GA-based layer. Furthermore, they used three types of mutations in the inner GE-based layer. To test their approach, the authors used multiple datasets including CIFAR-10, CIFAR-100, MNIST, Fashion-MNIST, SVHN and Rectangles. Similarly to the  works conducted by Miikulainen et al.~\cite{DBLP:journals/corr/MiikkulainenLMR17} and Suganuma et al.~\cite{ijcai2018-755}, Assun{\c{c}}{\~{a}}o et al. performed only 10 epochs to train the DNNs due to the high computational cost associated to this. The authors ran 10  runs and reported competitive results compared to state-of-the-art algorithms including using other automatic CNNs designing methods. It is interesting to note that when the authors computed the average fitness and the average number of hidden layers  across the entire population, they noticed that these two trends were  opposite. That is, as the fitness increases over time, the number of hidden layers decreases over time. This would suggest that these two metrics are in conflict when optimising CNNs architectures.

Sun et al.~\cite{DBLP:journals/tec/SunYY19} proposed the use of a population-based GA, of a fixed-length encoding, to evolve, by means of selection, crossover and mutation, unsupervised DNNs for learning meaningful representations for computer vision tasks. Their approach included two main parts (i) finding the optimal architectures in DNNs, the desirable initialisation of weights as well as activation functions, and (ii) fine-tuning all the parameter values in connection weights from the desirable initialisation. The first was primarily achieved by using an encoding, which was inspired by the work conducted by Zhang et al.~\cite{10.5555/1597538.1597629}, who captured all of the elements described in (i). As one gene represents on average 1,000 parameters in this  encoding,  the exploitation achieved by crossover is reduced. To overcome this problem, Sun et al. used backpropagation in Part (ii). By hand-crafting the various parts of their approach, the authors demonstrated how the local search adopted in Part (i) was necessary in order to achieve promising results.

Recently, Sun et al.~\cite{8712430} proposed a GA, named Evolving Deep CNNs, to automatically discover CNN  architectures and corresponding connection weight values. Inspired by the large computational resources reported in the work by Real et al.~\cite{10.5555/3305890.3305981} who used 250 high-end computers, Sun et al. aimed to tackle the use of this resource intensive setting. They proposed a cost-effective method to evaluate the fitness of the individuals in the population allowing them to execute 30 independent runs, as normally adopted by the EA community, in each of the 9 datasets used in their experiments. Moreover, they also used the normal evolutionary operators from EAs, including selection, mutation and crossover. The latter operator was not used in the studies carried out by Real et al.~\cite{10.5555/3305890.3305981}. This was a limitation in Real's work because crossover deals with exploiting the search space.  In Sun et al.'s approach each variable-length chromosome encodes the convolutional layer, the pooling layer and the full connection layer.  Because hundred of thousands connection weights may exist in one convolutional or full connection layer these should not be directly encoded into the chromosome. Thus, Sun et al.  used two statistical measures: the standard deviation and the average value of the connection weights. By doing so, they were able to efficiently evaluate each chromosome in the population. They evaluated each individual by using the classification error as well the number of connection weights. To mitigate the computational time required to evaluate the chromosomes along with the normally CNN  deep architectures the authors restricted the training to a small number of epochs ($\le$ 10). It is in the last epoch where the fitness is computed for each of the chromosomes in the population. Crossover was applied by using unit alignment, which basically groups parts of the same type, cross them over, and finally, use a restoring phase to put the units back in their corresponding positions in the chromosome. The authors reported highly encouraging results with many cases achieving better results compared to state-of-the-art algorithms in the datasets commonly used by the deep learning community. 

van Wyk and Bosman~\cite{8852417} described and evaluated their proposed neural architecture search (NAS) method to automate the process of finding an optimal CNN architecture for the task of arbitrary image restoration. Their work demonstrates the potential feasibility for performing NAS under significant memory and computational time constraints. The ImageNet64x64 dataset was chosen for evaluation. The training set was used for gradient-based optimisation of the NNs performance on unseen data. The authors find that the human-based configured architecture was heavily overparameterised while this was not the case with the evolved NN which performed the tasks with a significantly lower number of total parameters. Sun et al.~\cite{8742788} proposed an encoding strategy built on the state-of-the-art blocks namely ResNet and DenseNet. The authors used a variable length GA that allowed them to automatically evolve CNNs architectures of unrestricted depth. Sun et al. used selection, crossover and mutation to evolve candidate solutions. Given the nature of the variable length encoding used in their approach, the authors employed a repair mechanism that allowed them to produce valid CNNs. The authors used the CIFAR-10 and the CIFAR-100 datasets and compared their results against 9 manually designed methods, 4 semi-automatic methods and 5 automatic methods.  Interestingly their results outperformed all hand-crafted methods as well as all semi-automatic methods in terms of the classification error rate. 

Evolutionary Multi-objective Optimisation~\cite{1597059,CoelloCoello1999,Deb:2001:MOU:559152}, explained in  Section~\ref{subsection:moo}, has been hardly used in the automatic configuration of DNNs networks as well as in the optimisation of their hyperparameters. Works on the latter include the recent approach proposed by Kim et al.~\cite{Kim2017NEMON}, where the authors used speed and accuracy as two conflicting objectives to be optimised by means of EMO through the use of the Nondominated Sorting Genetic Algorithm II (NSGA-II)~\cite{Deb02afast}. The authors reported interesting results using three classification tasks, including the use of the MNIST, CIFAR-10 and the Drowsy Behaviour Recognition datasets. Inspired by the Kim et al.~\cite{Kim2017NEMON} study,  Lu et al.~\cite{10.1145/3321707.3321729} used the same EMO with the same conflicting objectives. It is worth noting that Lu et al.~\cite{10.1145/3321707.3321729} empirically tested multiple computational complexity metrics to measure speed including number of active nodes, number of active connections between nodes and floating-point operations (FLOPs), to mention some. Lu et al. indicated that the latter metric was more accurate and used it as a second conflicting objective for optimisation. Moreover, the authors used an ingenious bitstring encoding in their genetic algorithm which allowed them to use common and robust genetic operators normally adopted in GAs, including homogeneous crossover and bit-flip mutation (at most one change for each mutation operation). The authors tested their EMO approach with the CIFAR-10 and CIFAR-100 datasets achieving competitive results against state-of-the-art algorithms, including reinforcement learning-based approaches and human expert configurations.

 Wang et al.~\cite{DBLP:conf/ausai/WangSXZ18} explored the ability of differential evolution to automatically evolve the architecture and hyper-parameters of deep CNNs. The method called DECNN uses differential evolution where control of the evolution rate is managed by the a differential value. The DECNN evolves variable-length architectures for CNNs. An IP-based encoding strategy is implemented here to use a single IP address to represent one layer of a DNN. This IP address is then pushed into a sequence of interfaces corresponding to the same order as the layers in DNNs. Six of the MNIST datasets are used for benchmark testing and the DECNN performed very competitively with 12 state-of-the-art competitors over the six benchmarks. Mart\'in et al.~\cite{MARTIN2018180} proposed EvoDeep which is an EA designed to find the best architecture and optimise the necessary parameters to train a DNN. It uses a Finite-State Machine model in order to determine the possible transitions between different kind of layers, allowing EvoDeep to generate valid sequences of layers, where the output of one layer fits the input requirements of the next layer. It is tested on the benchmark MNIST datasets. The authors report that the high commputational resources required to train the DNN means that future work is needed to make the whole process more computationally efficient.

\subsection{Deep Learning Architecture: AutoEncoders}

 Suganuma et al.~\cite{DBLP:conf/icml/SuganumaOO18} used Cartesian Genetic Programming~\cite{Miller2011}, adopting an ES (1+$\lambda$) technique, and using selection and mutation operators only, to optimise DNS architectures for image restoration. To this end, the authors used convolutional autoencoders (CAEs) built upon convolutional layers and skip connections. By optimising the network in a confined search space of symmetric CAEs, the authors achieved competitive results against other methods without the need of using adversarial training and sophisticated loss functions, normally employed for image restoration tasks.
 
So et al.~\cite{DBLP:conf/icml/SoLL19} evolved a transformer network to be used in sequence-to-sequence language tasks. The encoding search space adopted by the authors was inspired by the NASNet search space proposed by Zoph et al.~\cite{DBLP:journals/corr/ZophVSL17}, see Fig.~\ref{fig:NASNet} (a). This was modified to express characteristics found in state-of-the-art feed-forward seq2seq networks such as the transformer network used in their work. The minimally tuned search space helped them to seed the initial population using a known transformer model. Because computing the fitness of the population, through the use of the negative log perplexity of the validation set, is time consuming, So et al. proposed Progressive Dynamic Hurdles. In essence, the latter allowed promising solutions to be trained using larger training datasets compared to poor potential solutions. The proposed mechanism, using selection and mutation operators only, yielded better results compared to transformer models in the language tasks used in their studies. Hajewski et al. \cite{hajewski2020distributed} describe an efficient and scalable EA for neural network architecture search with application to the evolution of deep encoders. Lander and Shang~\cite{7273701} introduce EvoAE an evolutionary algorithm for training autoencoders for DNNs. This proposed methodology is aimed at improving the training time of autoencoders for constructing DNNs. The EvoAE approach searches in the network weight and network structure space of the autoencoders simultaneously allowing for dual optimality searching. Large datasets are decomposed into smaller batches to improve performance. The good performance on large datasets distributed in cloud-based systems could be a major advantage of this approach as a population of high quality autoencoders can be created efficiently. 

Luo et al.~\cite{LUO2019103814Faults2019} propose a novel semi-supervised autoencoder called a discriminant autoencoder for application in fault diagnosis. Here, the proposed discriminant autoencoder has a different training process and loss function from traditional autoencoders. In the case of the discriminant autoencoder it is capable of extracting better representations from the raw data provided. A completely different loss function is used and the representations extracted by the discriminant autoencoder can generate bigger differences between the sample classes. The discriminant autoencoder makes full use of labels and feature variables to obtain the optimal representations, based on which the centers of the groups of samples that can be separated as much as possible. Ashfahani et al.~\cite{DEVDAN_ASHFAHANI2020297} propose DEVDAN as a deep evolving denoising autoencoder for application in data stream analytics. DEVDAN demonstrates a proposal of a denoising autoencoder which is a variant of the traditional autoencoder but focused on retrieving the original input information from a noise pertubation. DEVDAN features an open structure where it is capable of initiating its own structure from the beginning without the presence of a pre-configured network structure. DEVDAN can find competitive network architecture compared with state-of-the-art methods on the classification task using ten prominent datasets (including MNIST).

\subsection{Deep Learning Architecture: Deep Belief Networks}
\label{sec:architecture:dbn}

In the work by Chen et al.~\cite{7744376} the authors use DBNs to automatically extract features from images. They propose the EFACV (Evolutionary Function Array Classifier Voter) which classifies features from images extracted by a DBN (composed of stacked RBMs). An evolutionary strategy is used to train the EFACV and is mainly used for binary classification problems. For multi-class classification problems it is necessary to have multiple EFACV. The EFACV shows fast computational speed and a reduction in overall training time. Experiments are performed on the MNIST dataset. Liu et al.~\cite{7920404} describe structure learning for DNNs based on multiobjective optimisation. They propose a multiobjective optimisation evolutionary algorithm (MOEA). The DBN and its learning procedure use an RBM to pretain the DN layer by layer. It is necessary to remove unimportant or unncessary connections in DNN and move toward discovering optimal DNN connection structure which is as sparse as possible without lost of representation. Experiments based on the MNIST and CIFAR-10 datasets with different training samples indicate that the MOEA approach is effective.

Zhang et al.~\cite{7508982} use DBNs for a prognostic health management system in aircraft and aerospace design. DBNs offer a promising solution as they can learn powerful hierarchical feature representations from the data provided. The authors propose MODBNE (multiobjective deep belief networks ensemble) which is a powerful multiobjective evolutionary algorithm named MOEA based on decomposition. This is integrated into the training of DBNs to evolve multiple DBNs simultaneously with accuracy and diversity as two conflicting objectives in the problem. The DBN is composed of stacked RBMs which are trained in an unsupervised manner. MODBNE is evaluated and compared against a prominent diagnostics benchmarking problem with the NASA turbofan engine degradation problem. In the proposed approach the structural parameters of the DBN are strongly dependent on the complexity of the problem and the number of training samples available. The approach worked outstandingly well in compassion to other existing approaches. GPU-based implementations will be tested in the future for MODBNE to investigate the acceleration of computational processing speed.  Zhang et al.~\cite{DBLP:journals/tnn/0003T0H19} consider the problem of cost-sensitive learning methods. The idea of cost-sensitive learning is to assign misclassification costs for each class appropriately. While the authors report that there are very few studies on cost-sensitive DBNs these networks have drawn a lot of attention by researchers recently. Imbalances in the classes in input data is a problem. If there is a disproportionate number of class instances this can affect the quality of the applied learning algorithms. Zhang et al. argue that DBNs are very well placed to handled these type of imbalanced data problems. The ECS-DBN (Evolutionary Cost-Sensitive Deep Belief Network) is proposed to deal with such problems by assigning differential misclassification costs to the data classes. The ECS-DBN is evaluated on 58 popular Knowledge Extraction-based on Evolutionary Learning (KEEL) benchmark datasets. 

\begin{table*}[th!]
\caption{Summary of EA representations, genetic operators, parameters and its values used in neuroevolution in the design of DNNs architectures, along with the datasets used in various studies with their corresponding computational effort given in GPU days. Automatic and semi-auto(matic) refer to works where the architecture has been evolved automatically or by using a semi-automatic approach, such as using a constrained search space, respectively.
The dash (--) symbol indicates that the information was not reported or is not known to us.}

\centering
\resizebox{0.99\textwidth}{!}{ 
  \begin{tabular}{p{2.2cm}p{1.5cm}p{2cm}p{0.5cm}p{0.5cm}p{0.5cm}p{1.2cm}p{0.5cm}p{0.5cm}p{2.5cm}p{3cm}p{1.5cm}p{1.3cm}r}
\hline
Study                 & EA & Representation  &  \multicolumn{3}{c}{Genetic Operators}    & \multicolumn{3}{c}{EAs Parameters' values}&  Computational & Datasets  & GPU days   &   Automatic/  & DNNs\\
                      & Method       &                 & Cross & Mut & Selec & Pop & Gens & Runs                     &  Resources   &    & per run  &  Semi-auto & \\
\hline

Agapitos et al.~\cite{DBLP:conf/cec/AgapitosONFKBC15} & GP & Variable length & \cmark &\cmark & \cmark & 500, 1,000 & 100 & 30 & -- &   MNIST &-- &  Automatic & CNN\\ \hline 

 Assun{\c{c}}{\~{a}}o et al.~\cite{10.1007/978-3-319-77553-1_2,DBLP:journals/gpem/AssuncaoLMR19} & GAs, GE& Fixed and variable length& \cmark & \cmark & \cmark &100 &100 &10 &Titan X GPUs &CIFAR-10, 3 MNIST variants, Fashion, SVHN, Rectangles, CIFAR-100 &  -- &  Automatic & CNN \\ \hline 

 Charte et al.~\cite{CharteEvoAA2020} & GAs, ES, DE & Variable length & \cmark & \cmark & \cmark & $150$ & 20 & -- & 1 NVidia RTX 619 2080 GPU   &  CIFAR10, Delicious, Fashion, Glass, Ionosphere, MNIST, Semeion, Sonar, Spect & Limited to 24 hours& Automatic&AE   \\ \hline

 Chen et al.~\cite{7744376}& EAs & Fixed length & \xmark &\cmark & \cmark & 1 + $\lambda$, $\lambda$=4 & 15000 & $30$ & -- &  MNIST & -- & Automatic & DBN\\ \hline

 Desell~\cite{DBLP:conf/gecco/Desell17} & NEAT-based & Variable length & \cmark & \cmark &\cmark & 100 & & 10 & 4,500 volunteered PCs & MNIST& -- & Semi-auto & CNN \\ \hline

Goncalvez et al.~\cite{Goncalves2020} & GAs & Fixed length & \xmark & \cmark & \cmark & 1 + $\lambda$ $\lambda$ =4 & gens & runs & No GPUs & 4 Binary Class datasets: Cancer, Diabetes, Sonar, Credit & --& Automatic & CNN \\ \hline

Hajewski et al.~\cite{hajewski2020distributed} & EAs & Variable length & \cmark & \cmark & \cmark &  $\mu$ +$\lambda$ = 10 ($\lambda$,$\mu$ not specified & 20 & 20-40 & AWS ( Nvidia K80 GPU)& STL10 &--&Automatic&AE   \\ \hline

 Kim et al.~\cite{Kim2017NEMON} & EAs & -- & -- & -- & -- & 50, 40, 60 & -- & --& 60 Tesla M40 GPUs  & MNIST, CIFAR-10, Drowsiness Recognition & -- & Semi-auto & CNN \\ \hline

Lander et al.~\cite{7273701} & GAs & Variable length & \cmark & \cmark & \cmark & $30$ & $50$ & $5$ & No GPU &  MINST & - & Automatic & AE  \\ \hline

Liu et al.~\cite{7920404} & EAs & Variable length & \cmark & \cmark & \cmark & $100$ & $5000$ & $30$ & -- &  MINST, CIFAR-10 & - & Automatic & AE,RBM  \\ \hline

Lu et al.~\cite{10.1145/3321707.3321729} & GAs & Fixed length & \cmark & \cmark& \cmark&40  &30 & -- & 1 NVIDIA 1080Ti & CIFAR-10, CIFAR-100 & 8 in both & Automatic & CNN \\ \hline

 Mart\'in et al.~\cite{MARTIN2018180} evoDeep &EAs & Fixed Length & \cmark & \cmark & \cmark & $\lambda$ + $\mu$ = 10 ($\lambda$,$\mu$ =5) & 20 & -- & MNIST & Automatic & -- & CNN \\ \hline

 Peng et al.~\cite{PENG20181301}& EAs & Fixed length & \cmark &\cmark & \cmark & $10$ & $20$ & $-$ & -- & Electricity price & - & Automatic & LSTM  \\ \hline

Real et al.~\cite{DBLP:conf/aaai/RealAHL19} & GAs& Variable length &  \xmark & \cmark &\cmark    &100 & -- &5  &450 K40 GPUs &   CIFAR-10, ImageNet & 3150, 3150 & Semi-auto & CNN\\  \hline

 Real et al.~\cite{10.5555/3305890.3305981}  & GAs &   Variable length  &  \xmark    & \cmark & \cmark & 1000 &   -- & 5 & 250 PCs         & CIFAR-10, CIFAR-100  & 2750, 2750 & Automatic & CNN\\
\hline

 Shinozaki and Watanabe~\cite{7178918}&GAs, ES&Fixed length& \cmark &\cmark & \cmark&62&30&--&62 GPUs  &  AURORA2 Spoken Digits Corpus & 2.58 &   Automatic & RBM\\  \hline

  So et al.~\cite{DBLP:conf/icml/SoLL19} & EAs & Fixed length &\xmark & \cmark& \cmark& 100 &-- & --&200 workers with 1 Google TPU V.2 chip & WMT En-De, WMT En-Fr, WMT En-Cs &-- & Semi-auto & AE\\ \hline

  Suganuma et al.~\cite{DBLP:conf/icml/SuganumaOO18} & ES & Fixed length & \xmark &\cmark &\cmark &  1+$\lambda$, $\lambda=\{1,2,4,8,16\}$  & 250 & -- & 4 P100 GPUs &  Cars, CelebA, SVNH & 12 (inpainting), 16 (denoising) &  Automatic & AE \\ \hline

 Suganuma et al.~\cite{ijcai2018-755} & GP, ES  & Variable length & \xmark       & \cmark      & \cmark      & 1+$\lambda$, $\lambda=2$     & 300, 500, 1500    & 3   & Multiple PCs, 2 GPUs:  GTX 1080, Titan X&  CIFAR-10 (2 variants) &  27, 27   & Automatic & CNN\\
\hline

Sun et al.~\cite{8712430}    & GAs    & Variable length &  \cmark &  \cmark &  \cmark & 100 & 100 & 30 &  1 PC, 2 GTX 1080 GPUs &   Fashion, Rectangle (2 variants), Convex Set, MNIST (5 variants) & 8 (fashion), 5 (others) & Automatic&  CNN \\ \hline

 Sun et al.~\cite{8742788} & GAs & Variable length & \cmark & \cmark & \cmark &20 &20  &5 & 3 GTX 1080 Ti GPUs  & CIFAR-10, CIFAR-100 &  27, 36&  Automatic & CNN\\ \hline

 Sun et al.~\cite{8712430}  & GAs & Variable length&\cmark &\cmark &\cmark & 100 &100 & 30 & 2 GTX1080 GPUs & Fashion, Rectangle (2 variants), Convex Set, MNIST (5 variants)  & 8 (Fashion), 5 (others) & Automatic & CNN\\ \hline

 Sun et al.~\cite{DBLP:journals/tec/SunYY19} & GAs& Fixed length & \cmark &\cmark &\cmark & 50 & -- &30 & -- & Fashion, Rectangle (2 variants), Convex Set, MNIST (5 variants) CIFAR-10-BW & -- & Automatic & CNN\\ \hline

 van Wyk and Bosman~\cite{8852417} & EAs& Fixed length & \cmark &\cmark &\cmark & 20 & 20000 &1 & 1 GPU (GTX 1080) & ImageNet64x64 & Halted after 2 hrs & Automatic & CNN\\ \hline

Wang et al.~\cite{DBLP:conf/ausai/WangSXZ18} & EAs & Variable length & \cmark &\cmark & \cmark & $30$ & $20$ & $30$ & -- &  MNIST (5 variants) and Convex Set & --& Automatic & CNN \\ \hline

 Xie et al.~\cite{8237416} & GAs & Fixed length & \cmark & \cmark & \cmark & 20 &50 & --& 10 GPUs (type not specified) & CIFAR-10, MNIST, ILSVRC2012, SHVN &   17 (CIFAR-10), 2 (MNIST), 20 (ILSVRC2012), -- (SHNV)  &  Semi-auto & CNN \\ \hline

 Zhang et al.~\cite{7508982} & EAs & Variable length & \xmark &\cmark & \cmark & $20$ & $500$ & $10$ & No GPUs &   NASA C-MAPSS (Aircraft Engine Simulator Datasets)&-- & Automatic&  DBN\\ \hline

 Zhang et al.~\cite{DBLP:journals/tnn/0003T0H19} & EAs & Fixed Length & \cmark &\cmark & \cmark & -- & $30$ & 10 & 1 NVIDIA GTX 980 GPU    & 58 Knowledge Extraction based on Evolutionary Learning (KEEL) datasets &--& Automatic & DBN \\ 

\hline
\end{tabular}
}
\label{tab:eas}
\end{table*}

\subsection{Other networks: LSTM, RRN, RBM}

Shinozaki and Watanabe~\cite{7178918} proposes an optimisation strategy for DNN structure and parameters using an EA and a GA. The DNN structure is parameterised by a directed acyclic graph. Experiments are carried out on phoneme recognition and spoken digit detection. All of the experiments were conducted upon a massively parallel computing platform where the experiments were ran using 62 general-purpose computing on graphics processing units (GPGPUs). RBMs are used in the training phase. Ororbia et al.~\cite{10.1145/3321707.3321795} develop an evolutionary algorithm called EXAMM (Evolutionay eXploration of Augmenting Models) which is designed to devolve recurrent neural networks (RNNs) using a selection of memory structures. RNNs are particularly well suited to the task of performing prediction of large-scale real-world time series data. EXAMM was design to select from a large number of memory cell structures and this allowed the evolutionary approach to yield the best performing RNN architecture.

In Peng et al.~\cite{PENG20181301} the authors propose the LSTM (long short-term memory) neural network which is capable of analysing time series over long time spans in order to make predictions and effectively tackle the vanishing gradient problem. Their study uses differential evolution (DE) to identify the hyperparameters of the LSTM. DE approaches have been shown to out perform other approaches such as particle swarm optimisation and GAs. The authors claim that this is the first time that DE has been used to choose hyperparameters for LSTM for forecasting applications. As forecasting involves complex continuous nonlinear functions, the DE approach is well suited to these types of problems.  Goncalvez et al.~\cite{Goncalves2020} introduce a neuroevolution algorithm called the Semantic Learning Machine (SLM) which has been shown to outperform other similar methods in a wide range of supervised learning problems. SLM is described as a geometric semantic hill climber approach for NNs following a $1 + \lambda$ strategy. In the search for the best NN architecture configuration this allows the SLM to concentrate on the current best NN without drawing any penalties for this. The crucial aspect of the SLM approach is the geometric semantic mutation which takes a parent NN and generates a child NN.

\subsection{Final Comments}

The use of evolution-based method in designing deep neural networks is already a reality as discussed in this section. Different EAs methods with different representations have been used, ranging from landmark evolutionary methods including  Genetic Algorithms, Genetic Programming and Evolution Strategies up to using hybrids combining, for example, the use of Genetic Algorithms and Grammatical Evolution. In a short period of time, we have observed both ingenious representations and interesting approaches achieving extraordinary results against human-based design networks~\cite{DBLP:conf/aaai/RealAHL19} as well as state-of-the-art approaches, in some case employing hundred of computers~\cite{10.5555/3305890.3305981} to using just a few GPUs~\cite{8712430}. We have also learnt that most of the neuroevolution studies has focused their attention in designing deep CNNs. Other networks have also been considered including AE, RBM, RNN, LSTM and DBM, although there are just a few neuroevolution works considering the use of these types of networks. 


Table~\ref{tab:eas} contains extracted information from almost $30$ selected papers in neuroevolution. We selected these papers in our own ad-hoc way in order to find a selection of papers which succinctly demonstrated the use of neuroevolution in deep neural networks. The table is order in alphabetically order of the lead-author surname and summarises: the EA representation used, the representation of individuals, which genetic operators are used, and the EA parameters. The table also outlines the computational resources used in the corresponding study by attempting to outline the number of GPUs used. A calculation of the GPU days per run is approximated in the same way as Sun et al.~\cite{8742788}. The table states which benchmark datasets are used in the experimental analysis. Finally, the table indicates if the neural network architecture has been evolved automatically or by using a semi-automated approach whilst also indicating the target DNN architecture. Every selected paper does not report the same information which some papers omitting details about the computation resources used and others omitting information about the number of runs performed. One of the very interesting outputs from this table is that there are numerous differences between the approaches used by all of the papers listed in the table. We see crossover being omitted from several studies mostly due to encoding adopted by various researchers. Population size and selection strategies for the EAs change between studies. While MNIST and CIFAR are clearly the most popular benchmark datasets we can see many examples of studies using benchmark datasets from specific application domains.  

\section{Training Deep Neural Networks Through Evolutionary Algorithms}
\label{section:learningDNNsThroEAs}


\subsection{Motivation}

Backpropagation has been one of the most successful and dominant methods used in the training of ANNs over the past number of decades~\cite{10.5555/104279.104293}. This simple, effective and elegant method applies Stochastic Gradient Descent (SGD) to the weights of the ANN where the goal is to keep the overall error as low as possible.  However, as remarked by Morse and Stanley~\cite{10.1145/2908812.2908916} the widely held belief, up to around 2006, was that backpropagation would suffer loss of its gradient within DNNs. This turned out to be a false assumption and it has subsequently been proved that backpropagation and SGD are effective at optimising DNNs even when there are  millions of connections. Both backpropagation and SGD benefit from the availability of sufficient training data and the availability of computational power. In a problem space with so many dimensions the success of using SGD in DNNs is still surprising. Practically speaking, SGD should be highly susceptible to local optima~\cite{10.1145/2908812.2908916}. EAs perform very well in the presence of saddle points as was discussed in Section~\ref{sec:background}.

\subsection{The critique}
As there are no guarantees of convergence the solutions computed using EAs are usually classified as near optimal. Population-based EAs are in effect an approximation of the gradient as this is estimated from the individuals in a population and their corresponding objectives. On the other hand, SGD computes the exact gradient. As a result some researchers may consider EAs unsuitable for DL tasks for this reason. However, it has been demonstrated that the exact approximation obtained by SGD is not absolutely critical in the overall success of DNNs using this approach. Lillicrap et al.~\cite{LilCow14Random} demonstrated that breaking the precision of the gradient calculation has no negative or detrimental effect on learning. Indeed, Morse and Stanley~\cite{10.1145/2908812.2908916} speculated that the reason for the lack of research focus on using evolutionary computation in DNNs was not entirely related to concerns around the gradient. It more than likely resulted from the belief that new approaches to DNN could emerge from outside of SGD.

\subsection{Deep Learning Architecture: Convolutional Neural Networks}

Such et al.~\cite{Such2017DeepNG} proposed a gradient-free method to evolve the weights of convolutional DNNs by using a simple GA, with a population of chromosomes of fixed length. The proposed mechanism successfully evolved networks with over four million free parameters. Some key elements in the study conducted by Such et al. to successfully evolve these large neural networks include (i) the use of the selection and mutation genetic operators only (excluding the use of the crossover operator), (ii) the use of a novel method to store large parameter vectors compactly by representing each of these as an initialisation seed plus the list of the random seeds that produces the series of mutations that produced each parameter vector, (iii) the use of a state-of-the-art computational setting, including one modern computer with 4 GPUs and 48 CPU cores as well as 720 CPU cores across dozens of computers. Instead of using a reward-based optimisation techniques by means of a fitness function, Such et al. used novelty search~\cite{novelty2011} that rewards new behaviours of individuals. The authors used reinforcement learning benchmark problems including atari 2600~\cite{DBLP:journals/corr/abs-1207-4708,DBLP:journals/corr/MnihKSGAWR13}, hard maze~\cite{10.1162/EVCO_a_00025} and humanoid locomotion~\cite{brockman2016openai}. They demonstrated how their proposed approach is competitive with state-of-the-art algorithms in these problems including DQN~\cite{DBLP:journals/corr/MnihKSGAWR13}, policy-gradient methods~\cite{SEHNKE2010551} and ES~\cite{salimans2017evolution}. 

Pawelczyk et al.~\cite{10.1145/3205651.3208763} focused their attention in encoding CNNs with random weights using a GA, with the main goal to let the EA to learn the number of gradient learning iterations necessary to achieve a high accuracy error using the MNIST dataset. It was interesting to observe that their EA-based approach reported the best results  with around 450 gradient learnt iterations compared to 400 constant iterations which yielded the best overall results.

\subsection{Deep Learning Architecture: Autoencoders}
David and Greental~\cite{10.1145/2598394.2602287} used a GA of fixed length to evolve the weight values of an autoencoder DNN. Each chromosome was evaluated by using the root mean squared error for the training samples. In their experiments, the authors used only 10 individuals with a 50\% elitism-policy. The weights of these individuals were updated using backpropagation and the other half of the population were randomly generated in each generation. They tested their approach with the well-known CIFAR-10 dataset. They compared their approach \emph{vs.} the traditional autoencoder using SVM, reporting  a better classification error when using their proposed GA-assisted method for the autocoder DNN  (1.44\% \emph{vs.} 1.85\%). In their studies, the authors indicated that the reason why their method produced better results was because gradient descent methods such as backpropagation are highly susceptible to being trapped at local optima and their GA method helped to prevent this.  

Fernando et al.~\cite{10.1145/2908812.2908890} introduced a differentiable version of the Compositional Pattern Producing Network (CPPN) called the Differentiable Pattern Producing Network (DPPN). The DPPN approach attemps to combine the advantages and results of gradient-based learning in NN with the optimisation capabilities of evolutionary approaches. The DPPN has demonstrated superior results for the benchmark dataset MNIST. A generic evolutionary algorithm is used in the optimisation algorithm of DPPN. The results indicate that the DPPNs and their associated learning algorithms have the ability to dramatically reduce the number of parameters of larger neural networks. The authors argue that this integration of evolutionary and gradient-based learning allows the optimisation to avoid becoming stuck in local optima points or saddle points.

\subsection{Other Relevant Works}
Morse and Stanley~\cite{10.1145/2908812.2908916} proposed an approached called limited evaluation evolutionary algorithm (LEEA), that effectively used a population-based GA of fixed length representation to evolve, by means of crossover and mutation, 1000 weights of a fixed-architecture  network. The authors took inspiration from SGD that can compute an error gradient from a single (or small batch of) instance of the training set.  Thus, instead of computing the fitness of each individual in the population using the whole training set, the fitness is computed using a small fraction. This results in an EA that computationally similar to SGD. However, using such approach is also one of the weakness in LEAA because it does not generalise to whole training sample. To mitigate this, the authors proposed two approaches: (i) the use of a small batch of instances and (ii) the use of a fitness function that consider both the performance on the current mini-batch and the performance of individuals' ancestors against their mini-batches.  To test their idea, the authors used a function approximation task, a time series prediction task and a house price prediction task and compared the results yield by their approach against SGD and RMSProp. They showed how their LEEA approach was competitive against the other approaches. Even when the authors do not use DNNs, but a small artificial NN, it is interesting to note how this can be used in a DNN setting.

Khadka and Tumer~\cite{Khadka10.5555/3326943.3327053} remark that Deep Reinforcement Learning methods are ``notoriously sensitive to the choice of their hyperparamaters and often have brittle convergence properties''. These methods are also challenged by long time horizons where there are sparse rewards. EAs can respond very positively to these challenges where the use of fitness metrics allows EAs to tolerate the sparse reward distribution and endure long time horizons. However, EAs can struggle to perform well when optimisation of a large number of parameters is required. The authors introduce their Evolutionary Reinforcement Learning (ERL) algorithm. The EA is used to evolve diverse experiences to train an RL agent. These agents are then subjected to mutation and crossover operators to create the next generation of agents. From the results outlined in the paper this ERL can be described as a ``population-driven guide'' that guides or biases exploration towards states with higher and better long-term returns, promoting diversity of explored policies, and introduces redundancies for stability.  

Recurrent Neural Networks (RNNs) (see Section~\ref{subsection:othernetworks}) incorporate memory into an NN by storing information from the past within the hidden states network. In~\cite{DBLP:journals/ec/KhadkaCT19}, Kahdka et al. introduce a new memory-augmented network architecture called the Modular Memory Unit (MMU). This MMU disconnects the memory and central computation operations without requiring costly memory management strategies. Neuroevolutionary methods are used to train the MMU architecture. The performance of the MMU approach with both gradient descent and neuroevolution are examined in the paper. The authors find that neuroevolution is more repeatable and generalizable across tasks. The MMU NN is designed to be highly configurable and this characteristic is exploited by the the neuroevolutionary algorithm to evolve the network. Population size is set to $100$ with a fraction of elites set at $0.1$. In the fully differentiable version of the MMU gradient descent performs better for Sequence Recall tasks than neuroevolution. However, neuroevolution performs significantly better than gradient descent in Sequence Classification tasks.

\begin{table*}[th!]
\caption{Summary on EA representations, genetic operators, parameters and its values used in neuroevolution in the training of DNNs, along with the datasets used in various studies with their corresponding computational effort given in GPU days. The dash (--) symbol indicates that the information was not reported or is not known to us.}
\centering
\resizebox{0.99\textwidth}{!}{ 
  \begin{tabular}{p{2.2cm}p{1.5cm}p{2cm}p{0.5cm}p{0.5cm}p{0.5cm}p{1.3cm}p{0.5cm}p{0.7cm}p{2.5cm}p{3cm}p{2cm}r}
\hline
Study                 & EA & Representation  &  \multicolumn{3}{c}{Genetic Operators}    & \multicolumn{3}{c}{EAs Parameters' values}&  Computational & Datasets  & GPU days    & DNN\\
                      & Method       &                 & Cross & Mut & Selec & Pop & Gens & Runs                     &  Resources   &    &per run   & \\
\hline

David and Greental~\cite{10.1145/2598394.2602287} & GAs & Fixed length & \cmark& \cmark& \cmark &10 & & --& -- &  MNIST  & -- & AE \\ \hline

 Dufourq and Bassett~\cite{8261132} & GAs & Variable length & \xmark & \cmark  & \cmark & 100 & 10 & 5 & 1 GTX1070 GPU  &  CIFAR-10, MNIST, EMNIST (Balanced \& Digits), Fashion, IMDB, Electronics  &--  & CNN\\ \hline

Fernando et al.~\cite{10.1145/2908812.2908890} & GAs &-- &\cmark &\cmark & \cmark & 50& --& --&-- & MNIST, Omniglot & -- & AE\\ \hline

 Khadka and Tumer~\cite{Khadka10.5555/3326943.3327053} & EAs & Variable length & \cmark & \cmark & \cmark & 10 & $\infty$ & 5& -- & 6 Mujoco (continuous control) datasets  & -- & Read text \\ \hline

 Khadka et al.~\cite{DBLP:journals/ec/KhadkaCT19} & EAs & Fixed length &\xmark  & \cmark & \cmark & 100 & 1000 10000 15000 &-- & GPU used but not specified & Sequence Recall, Sequence Classification  & -- & Read text \\ \hline

Morse and Stanley~\cite{10.1145/2908812.2908916} & GAs& Fixed length & \cmark & \cmark &\cmark &1,000 &-- & 10 & --& Function Approximation, Time Series, California Housing&-- & Read text\\ \hline

 Pawelczyk et al.~\cite{10.1145/3205651.3208763} & GAs & Fixed length & \cmark & \cmark & \cmark & 10 & -- & -- & 1 GPU (Intel Core i7 7800X, 64GB RAM) &   MNIST & --  & CNN\\ \hline

Such et al.~\cite{Such2017DeepNG} & GAs& Fixed length & \xmark & \cmark &\cmark & 1,000 (\underline{A}), 12,500 (\underline{H}), 20,000 (\underline{I})& --& 5 (\underline{A}), 10 (\underline{I})  & 1 PC (4  GPUs, 48  CPUs) and 720  CPUs  across  dozens  of PCs & \underline{A}tari 2600, \underline{I}mage Hard maze, \underline{H}umanoid locomotion & 0.6 (Atari, 1 PC), 0.16 (Atari, dozens of PCs) & CNN\\ 

\hline
\end{tabular}
}
\label{tab:learning}
\end{table*}

\subsection{Final Comments}

In the early years of neuroevolution, it was thought that evolution-based methods might exceed the capabilities of backpropagation~\cite{784219}. As ANNs, in general, and  as DNNs, in particular, increasingly adopted the use of stochastic gradient descent and backpropagation, the idea of using EAs for training DNNs instead has been almost abandoned by the DNN research community. EAs are a ``genuinely different paradigm for specifying a search problem''~\cite{10.1145/2908812.2908916} and provide exciting opportunities for learning in DNNs. When comparing neuroevolutionary approaches to other approaches such as gradient descent, authors such as Khadka et al.~\cite{DBLP:journals/ec/KhadkaCT19} urge caution. A generation in neuroevolution is not readily comparable to a gradient descent epoch. 

Despite the fact that it has been argued that EAs can compete with gradient-based search in small problems as well as using neural networks with a non-differentiable activation function~\cite{MANDISCHER200287}, the encouraging results achieved in the 1990s~\cite{Goerick_evolutionstrategies:,10.5555/1623755.1623876,10.1109/64.393138} have inspired recently some researchers to carry out research in training DNNs including the works conducted by David and Greental~\cite{10.1145/2598394.2602287} and Fernando et al.~\cite{10.1145/2908812.2908890} both works using deep AE as well as the works carried out by Pawelczyk et al.~\cite{10.1145/3205651.3208763} and
Such et al.~\cite{Such2017DeepNG}, both studies using deep CNNs. 

Table~\ref{tab:learning} is structured in a similar way to Table~\ref{tab:eas}. As with Table~\ref{tab:eas}, we selected these papers in our own ad-hoc way in order to find a selection of papers which succinctly demonstrated the use of EAs in the training of DNNs. As before we see mutation and selection used by all of the selected works with crossover omitted in certain situations. We see greater diversity in the types of benchmark datasets used with a greater focus on domain-specific datasets and problems. 


\section{Future Work on Neuroevolution in Deep Neural Networks}
\label{sec:challenges}

\subsection{Surrogate-assisted EAs in DNNs}

EAs have successfully been used in automatically designing artificial DNNs, as described throughout the paper, and multiple state-of-the-art algorithms have been proposed in recent years including  genetic CNN~\cite{8237416}, large-scale evolution~\cite{10.5555/3305890.3305981}, evolving deep CNN~\cite{8712430}, hierarchical evolution~\cite{DBLP:conf/iclr/LiuSVFK18}, to mention but a few successful examples. Despite their success in automatically configuring DNNs architectures, a common limitation in all these methods is the training time needed, ranging from days to weeks in order to achieve competitive results. Surrogate-assisted, or meta-model based, evolutionary computation uses efficient models, also known as meta-models or surrogates, for estimating the fitness values in evolutionary algorithms~\cite{JIN201161}. Hence, a well-posed surrogate-assisted EC considerably speeds up the evolutionary search  by reducing the number of fitness evaluations while at the same time correctly estimating the fitness values of some potential solutions.

The adoption of this surrogate-assisted EA is limited in the research discussed in this paper and is dealt with in a few limited exceptions. For example, in a recent work, Sun et al.~\cite{8744404} demonstrated how meta-models, using ensemble members, can be successfully used to correctly estimate the accuracy of CNNs. They were able to considerably reduce the training time, e.g., from 33 GPU days to 10 GPU days, while still achieving competitive accuracy results compared to state-of-the-art algorithms. A limitation in Sun et al.'s approach is the unknown number of training runs that is necessary to achieve a good prediction performance.

\subsection{Mutations and the neutral theory}

We have seen that numerous studies have used selection and mutation only to drive evolution in automatically finding a suitable DNN architecture (Section~\ref{sec:architectures}) or to train a DNN (Section~\ref{section:learningDNNsThroEAs}). Tables~\ref{tab:eas} and~\ref{tab:learning} present a summary of the genetic operators used by various researchers.  Interestingly, a good number of researchers have reported highly encouraging results when using these two genetic operators, including the works conducted by Real et al.~\cite{DBLP:conf/aaai/RealAHL19,10.5555/3305890.3305981} using  GAs and hundreds of GPUs as well as the work carried out by Suganuma et al.~\cite{ijcai2018-755} employing Cartesian Genetic Programming and a using a few GPUs. 

Kimura's neutral theory of molecular evolution~\cite{Kimura,kimura_1983} states that the majority of evolutionary changes at molecular level are the result of random fixation of \textit{selectively neutral mutations}. A mutation from one gene to another is neutral if it does not affect the phenotype. Thus, most mutations that take place in natural evolution are neither advantageous nor disadvantageous for the survival of individuals. It is then reasonable to extrapolate that, if this is how evolution has managed to produce the amazing complexity and adaptations seen in nature, then  neutrality should aid also EAs. However, whether neutrality helps or hinders the search in EAs is ill-posed and cannot be answered in general: one can only answer this question within the context of a specific class of problems, (neutral) representation and set of operators~\cite{DBLP:phd/ethos/GalvanLopez09,DBLP:conf/eurogp/LopezDP08,DBLP:conf/gecco/LopezP06,DBLP:conf/ppsn/LopezP06_2,DBLP:conf/micai/LopezP09,DBLP:journals/evs/LopezPKOB11,10.1007/978-3-540-73482-6_9,DBLP:journals/tec/PoliL12}. We are not aware of any works in neuroevolution in DNN on neutrality, but there are some interesting encodings adopted by researchers including Suganuma's work~\cite{ijcai2018-755} (see Fig.~\ref{fig:cartesianCNN}) that allow the measurement of the level of neutrality present in evolutionary search and indicate whether its presence is beneficial or not in certain problems and DNNs. If neutrality is beneficial, taking into consideration specific class of problems, representations and genetic operators, this can also have an immediate positive impact in the training time needed because the evaluation of potential EA candidate solutions will not be necessary. 

\subsection{Multi-objective Optimisation}
\label{subsection:moo}


The vast majority of works reviewed in this paper have focused their attention in the direct or indirect optimisation of one objective only. For example, when training a CNN in a computer vision supervised classification task, the classification error is normally adopted as a metric of performance for this type of network. Perhaps, one of the reasons why taking into account one objective has been the norm in the specialised literature is because the optimisation of one objective has been enough to yield extraordinary results (for example in the application domain of route optimisation~\cite{RoutePlanningBast2016}). Another potential reason could be due to the fact that  two or more objectives can be conflicting with each other making the (optimisation) task very difficult to  accomplished~\cite{1597059,CoelloCoello1999,Deb:2001:MOU:559152}. 

Multi-objective optimisation (MO) is concerned with the simultaneous optimisation of more than one objective function. When such functions are in conflict, a set of trade-off solutions among the objectives is sought as no single global optimum exists. The optimal trade-offs are those solutions for which no objective can be further improved without degrading one of the others. This idea is captured in the Pareto dominance relation: a solution $x$ in the search space is said to  {\em Pareto-dominate} another solution $y$ if $x$ is at least as good as $y$ on all objectives and strictly better on at least one objective. This is an important aspect in EMO (Evolutionary MO)~\cite{1597059,CoelloCoello1999,Deb:2001:MOU:559152}  because it allows solutions to be ranked according to their performance on all objectives with respect to all solutions in the population. EMO is one of the most active research areas in EAs. Yet it is surprising to see that EMO approaches have been scarcely used for the automatic configuration of artificial DNNs architectures or learning in DNNs. Often, the configuration of these artificial DNNs require simultaneously satisfying multiple objectives such as reducing the computational calculation of these on the training dataset while attaining high accuracy. EMO offers an elegant and efficient framework to handle these conflicting objectives. We are aware of only a few works in the area  e.g.,~\cite{Kim2017NEMON,7920404,10.1145/3321707.3321729,7508982}, as summarised in Section~\ref{sec:architectures}.

\subsection{Fitness Landscape Analysis of DNNs and Well-posed Genetic Operators}

As we have seen throughout the paper, all of the works in neuroevolution in DNNs have used core genetic operators including selection and mutation. Crossover has also been used in most of these works. The use of these operators are summarised in Tables~\ref{tab:eas} and~\ref{tab:learning}. The use of crossover, sometimes referred as recombination, can sometimes be difficult to adopt depending on the encoding used and some variants have been proposed such as in the study carried out in~\cite{8712430}. Other studies have adopted standard crossover operators such as those discussed in~\cite{Kim2017NEMON}. There are, however,  no works carried out in the area of neuroevolution in DNNs that have focused their attention in explaining why the adoption of a particular genetic operator is well-suited for that particular problem.

The notion of fitness landscape~\cite{Wright} has been with us for several decades. It is a non-mathematical aid that has proven to be very powerful in understanding evolutionary search. Viewing the search space, defined by the set of all potential solutions, as as landscape, a heuristic algorithm such as an EA, can be thought of as navigating through it to find the best solution (essentially the highest peak in the landscape). The height of a point in this search space, represents in an abstract way,  the fitness of the solution associated with that point. The landscape is therefore a knowledge interface between the problem and the heuristic-based EA. This can help researchers and practitioners to define well-behaved genetic operators, including mutation and crossover, over the connectivity structure of the landscape.

\subsection{Standardised Scientific Neuroevolution Studies in DNNs}

As described in Section~\ref{sec:background}, multiple DNNs architectures have been proposed in the specialised literature including CNNs, DBNs, RBMs and AEs.  Each of these DNNs considers multiple elements such as the activation function, type of learning, to mention a few examples. As indicated previously, EAs are incredible flexible allowing researchers to use elements from two or more different EAs methods. Moreover, multiple variants from each of these elements exists such as having multiple options from where to chose to exploit and explore the search space. Many of the research works reviewed in this paper have compared their results with those yield by neuroevolution-based state-of-the-art algorithms. However, it is unclear why some techniques are better than others. Is it because of the type of operators used? Is it because of the representation adopted in these studies  or is it because of the type of learning employed during training?  Due to the lack of standardised studies in neuroevolution on DNNs, it is difficult to draw final conclusions that help us to identify what elements are promising in DNNs.

\subsection{Diversifying the use of benchmark problems and DNNs}

\begin{table}
\caption{Common datasets used in neuroevolution in  deep  neural networks.}
\centering
\resizebox{1.00\columnwidth}{!}{ 
\begin{tabular}{lrrccc}
\hline
Data set         &  \multicolumn{2}{c}{Number of examples} & Input & RGB, B\&W, & No. of \\
                 & Training  & Testing                      & Size  & Grayscale  & classes\\ \hline
 MNIST~\cite{726791}    & 60,000   & 10,000 & 28$\times$28 & Grayscale  & 10\\
MNIST variants~\cite{10.1145/1273496.1273556} &  12,000 & 50,000 &  28$\times$28 & Grayscale & 10\\
CIFAR-10~\cite{cifar} & 50,000 &  10,000 & 32$\times$32 & RGB & 10\\
CIFAR-100~\cite{cifar} & 50,000   &10,000 & 32$\times$32 &  RGB & 100\\
Fashion~\cite{DBLP:journals/corr/abs-1708-07747}  & 60,000  & 10,000 & 28$\times$28 & Grayscale & 10 \\
SVHN~\cite{netzer_2011} & 73,257   & 26,032 & 32$\times$32 & RGB & 10 \\ 

Rectangle~\cite{10.1145/1273496.1273556} & 1,000         & 50,000 & 28 $\times$ 28 & B\&W & 2\\ 
Rectangle images~\cite{10.1145/1273496.1273556} & 10,000 & 50,000 & 28 $\times$28 & Grayscale & 2\\ 
Convex set~\cite{10.1145/1273496.1273556}& 6,000 &  50,000 & 28 $\times$ 28 & B\&W & 2\\ 
ILSVRC2012~\cite{5206848} & 1.3M &  150,000& 224 $\times$ 224 & RGB & 1,000 \\
GERMAN Traffic Sign Recognition~\cite{TrafficSignsSTALLKAMP2012323}&50,000&12,500&32$\times$32&Grayscale&43\\
CelebFaces~\cite{7346495}  & -- & -- & 39 $\times$ 31 & RGB & 2 \\
 (No. of images of CelebFaces: 87,628) & \\ \hline
  \multicolumn{6}{l}{\begin{minipage}{1.5\columnwidth}Number of examples for the validation set is omitted from this table given that it is well-known in the ML community that this is randomly split from the training data with the proportion of $\frac{1}{5}$.\end{minipage}}
\end{tabular}
}
\label{tab:computer_vision_datasets}
\end{table}

There is little argument that the availability of new and large datasets combined with ever increasingly powerful computational resources have allowed DNNs to tackle and solve hard problems in domains such as image classification, speech processing and many others. Image classification is certainly considered as the primary benchmark against which to evaluate DNNs~\cite{Zhu8573476Benchmarks}. These benchmark datasets (many of which are outlined in Table~\ref{tab:computer_vision_datasets}) are used as a means of comparing the computational results of experimental setups created by different research groups. We believe that the success of DNNs coupled with the need to tackle complex problems in other domains sees a growing need for DNNs to expand to other domains. In order to assess how successful DNNs are in other domains and with other practical problems robust and comprehensive benchmark datasets will be required. Indeed we believe that without such benchmarks it may be difficult to make convincing arguments for the success and suitability of DNNs for problems in other domains beyond image classification, machine translation, and problems involving object recognition. 

It is critical that benchmark datasets are available freely and as open-data. Stallkamp et al.~\cite{TrafficSignsSTALLKAMP2012323} argue that in a niche area such as traffic sign recognition it can be difficult to compare published work because studies are based on different data or consider classification in different ways. The use of proprietary data in some cases, which is not publicly available, makes comparison of results difficult. Authors such as Zhang et al.~\cite{7508982} access data from a prognostic benchmarking problem related to NASA and Aero-Propulsion systems. Specific problem domains outside of those of vision, speech recognition and language also have benchmark datasets available but may be less well-known. Zhang et al.~\cite{DBLP:journals/tnn/0003T0H19} use datasets from KEEL (Knowledge Extraction based on Evolutionary Learning) but also use a real-world dataset from a manufacturing drilling machine in order to obtain a practical evaluation. Chen and Li~\cite{ChenDeepLearningChallenges6817512} comment that as we see data getting bigger (so called Big Data) deep learning will continue to play an ever increasingly important role in providing big data predictive analytics solutions, particularly with the availability of increased processing power and the advances in graphics processors. However, while the potential of Big Data is without doubt, new ways of thinking and novel algorithmic approaches will be required to deal with the technical challenges.  Algorithms that can learn from massive amounts of data are needed ~\cite{ChenDeepLearningChallenges6817512} and this may make it difficult to define benchmark datasets within the Big Data domain.


\section{Conclusions}
\label{sec:conclusions}

This paper has provided a comprehensive survey of neuroevolution approaches in Deep Neural Networks (DNNs) and has discussed the most important aspects of application of Evolutionary Algorithms (EAs) in deep learning. The target audience of this paper is a broad spectrum of researchers and practitioners from both the Evolutionary Computation and Deep Learning (DL) communities. The paper highlights where EAs are  being used in DL and how DL is benefiting from this. Readers with a background in EAs will find this survey very useful in determining the state-of-the-art in neural architecture search methods in general. Additionally, readers from the DL community will be encouraged to consider the application of EAs approaches in their DNN work. Configuration of DNNs is not a trivial problem. Poorly or incorrectly configured networks can lead to the failure or under-utilisation of DNNs for many problems and applications. Finding well-performing architectures is often a very tedious and error-prone process. EAs have been shown to be a competitive and successful means of automatically creating and configuring such networks. Consequently, neuroevolution has great potential to provide a strong and robust toolkit for the DL community in future. The article has also outlined and discussed important issues and challenges in this area.

\bibliographystyle{abbrv}
\bibliography{neuroevolution}

\end{document}